\documentclass[11pt]{article}

\usepackage[margin=1in]{geometry}
\usepackage{amsmath, amssymb}
\usepackage{booktabs}
\usepackage{array}
\usepackage{graphicx}
\usepackage{float}
\usepackage{hyperref}
\usepackage{microtype}
\usepackage{parskip}
\usepackage{authblk}
\usepackage{algorithm}
\usepackage{algpseudocode}
\usepackage{xcolor}
\hypersetup{colorlinks=true, linkcolor=blue, citecolor=blue, urlcolor=blue}

\title{\textbf{When Latent Geometry Is Not Enough:\\
Draft-Conditioned Latent Refinement for Non-Autoregressive Text Generation}}

\author{De Shuai Zhang}
\affil{Beijing Wuzi University}
\date{Technical Report v1, May 2026 \\ \small\textit{Stage 1 (latent autoencoder and DraftPrior diagnostics) complete.}}

\begin{document}
\maketitle

\begin{abstract}
Continuous diffusion and flow models are attractive for non-autoregressive text generation because they can update all positions in parallel. A major difficulty is the interface between continuous latent states and discrete tokens. This report studies a draft-conditioned latent refinement model built from a frozen BERT encoder, a parallel decoder, a denoising DraftPrior, a local FlowNet, and a learned diagonal MetricNet. Early Gaussian-start experiments showed that good latent-space metrics, such as scale matching or cosine similarity, do not guarantee good decoding. Generated latents can be close to real encoder latents but still produce high-entropy, biased, or repetitive token distributions. We therefore frame the task as controlled local refinement rather than full generation from noise. On ROCStories, using the first two sentences as prompt and the last three as target, full 768-dimensional BERT latents recover tokens much better than compressed 256-dimensional latents. With 768-dimensional latents, DraftPrior target-token probability is 0.938 for clean drafts, 0.613 for 3\% token dropout, 0.483 for 5\% dropout, and 0.272 for 10\% dropout. Local flow refinement and fused decoder-aware readout give modest additional gains, while metric learning and OT-style alignment improve geometry but do not close the decoder gap. The main result is a diagnostic one: latent geometry alone is not enough. Continuous latent text generation should be evaluated by decoder recoverability, the quality of the start distribution, and whether refinement preserves decoder-readable structure.
\end{abstract}


\section{Introduction}

Autoregressive language models based on Transformer architectures~\cite{vaswani2017attention}, including GPT-style systems~\cite{radford2019language}, generate one token at a time. This factorization is powerful, but it makes generation sequential. Non-autoregressive generation is attractive because it could produce an entire continuation in parallel and permit global sequence-level refinement. The difficulty is that language is discrete, while diffusion and flow-based methods operate naturally in continuous spaces.

A common approach, related to denoising autoencoding models such as BART~\cite{lewis2020bart}, is to encode text into a continuous latent sequence, generate or refine in that latent space, and decode the resulting latents back into tokens. This resembles latent diffusion in vision, but the text setting is less tolerant. In images, a slightly imperfect latent may still decode into a plausible image. In text, a small latent error can change the token identity completely or leave the decoder uncertain across many vocabulary items.

This project began as a conditional latent flow-matching system. 
A frozen BERT encoder~\cite{devlin2019bert} maps an input token 
sequence $x = (x_1, \ldots, x_n)$ of length $n$ into continuous 
latent representations through a projection layer $P$:
\begin{equation}
    z = P(E_{\text{BERT}}(x)), \quad z \in \mathbb{R}^{n \times d}
\end{equation}
where $d$ is the latent dimension ($d = 768$ in our main 
experiments, $d = 256$ for the compressed baseline).
A parallel decoder $p_\phi$ predicts all token distributions 
simultaneously, forming a stage-1 latent autoencoder trained with:
\begin{equation}
    \mathcal{L}_{\text{AE}} = -\sum_{i=1}^{n} \log p_\phi(x_i \mid z)
\end{equation}
For conditional generation, the latent sequence is split into 
prompt and suffix parts at position $m$:
\begin{equation}
    z = (z^p, z^s), \quad z^p = z_{1:m}, \quad z^s = z_{m+1:n}
\end{equation}
where $z^p \in \mathbb{R}^{m \times d}$ is the known prompt latent 
and $z^s \in \mathbb{R}^{(n-m) \times d}$ is the suffix latent to 
be generated. In our ROCStories experiments, $m = 32$ prompt token 
slots and $n - m = 32$ suffix token slots.

Diagnostics showed that the original Gaussian-start formulation was 
too optimistic. A flow path interpolates between noise and the real 
suffix latent:
\begin{equation}
    z_0 \sim \mathcal{N}(0, I), \quad
    z_t = (1-t)z_0 + tz^s, \quad 
    t \sim \mathcal{U}(0,1)
\end{equation}
where $I$ is the identity matrix and $t$ is the interpolation 
timestep. A FlowNet $v_\theta$ is trained to predict the target 
velocity $u_t = z^s - z_0$:
\begin{equation}
    \mathcal{L}_{\text{FM}} = \mathbb{E}\left[
    \|v_\theta(z_t, t, z^p) - (z^s - z_0)\|^2\right]
\end{equation}
The flow could improve latent-space metrics while the decoder still 
produced high-frequency-token attractors, repetition, and poor 
target-token probability. Specifically, generated latents 
$\hat{z}^s$ could achieve high cosine similarity with real suffix 
latents while still decoding poorly:
\begin{equation}
    \cos(z^s, \hat{z}^s) \approx 1 \not\Rightarrow 
    P_{\text{target}}(\hat{z}^s) \text{ is high}
\end{equation}
where $P_{\text{target}}$ is the average target-token probability under the frozen decoder, defined formally in Eq.~\ref{eq:ptarget}.

The current version reframes the system. Instead of asking a flow 
to create a valid text continuation from pure noise, we use a rough 
draft as an instance-level guide. The draft is encoded into the 
same latent space, a DraftPrior maps it toward the decoder-readable 
basin, and the metric-aware flow applies only a small local 
correction. This report studies that controlled refinement setting 
and the failure modes that remain.
\subsection{Contributions}

This report makes four contributions:
\begin{itemize}
  \item We identify a decoder recoverability failure mode: latent similarity does not guarantee token recovery.
  \item We introduce a draft-conditioned refinement setup for ROCStories, using the first two sentences as prompt and the last three as target.
  \item We compare 256-dimensional compressed latents with full 768-dimensional BERT latents and show that stronger compression loses token-level information.
  \item We ablate DraftPrior, local FlowNet refinement, MetricNet, OT-style alignment, fused readout, and bounded residual refinement. The start distribution gives most of the recoverability; geometry-based refinement gives smaller gains.
\end{itemize}

\section{Related Work}

\subsection{Continuous diffusion and flow models for text}

Diffusion-LM~\cite{li2022diffusionlm}, building on diffusion 
models such as DDPM~\cite{ho2020ddpm} and score-based generative 
modeling~\cite{song2021scorebased} --- which learns the score 
function $s_\theta(x_t, t) \approx \nabla_{x_t} \log p(x_t)$ and 
generates samples via Langevin dynamics --- demonstrated that 
continuous diffusion can be adapted to text by mapping words into 
continuous embeddings and then rounding back to tokens. Its 
treatment of embedding choice and rounding is directly relevant 
here: the continuous-to-discrete interface is not an implementation 
detail, but a central modeling problem. In this report, the 
rounding operation is replaced by a parallel decoder, but the same 
challenge remains: generated continuous states must land in regions 
that the decoder can reliably map to tokens.

Diffusion-LM addresses the continuous-to-discrete gap with a 
clamping trick: during sampling, the predicted clean vector 
$f_\theta(\mathbf{x}_t,t)$ is mapped to its nearest word-embedding 
sequence before forming the next denoising state,
\begin{equation}
    \mathbf{x}_{t-1}
    =
    \sqrt{\bar{\alpha}_t}\,\mathrm{Clamp}(f_\theta(\mathbf{x}_t,t))
    +
    \sqrt{1-\bar{\alpha}_t}\,\epsilon.
\end{equation}
Our system faces the same interface problem, but uses a different 
constraint. Rather than snapping intermediate states to discrete 
word embeddings, we start from a decoder-readable draft latent and 
restrict the refinement magnitude:
\begin{equation}
    \mathbf{z}_{\mathrm{res}}
    =
    \mathbf{z}_{\mathrm{ode}}
    +
    \lambda\tanh R_\omega(\mathbf{z}_{\mathrm{ode}},\mathbf{z}^p)
\end{equation}
where $\lambda > 0$ is a scalar bound controlling the maximum 
refinement magnitude, ensuring
$\|\mathbf{z}_{\mathrm{res}} - \mathbf{z}_{\mathrm{ode}}\|_\infty 
\leq \lambda$.
Thus, Diffusion-LM enforces discreteness by repeated projection to 
the embedding vocabulary, while our bounded residual preserves 
local decoder-readability by preventing the refinement trajectory 
from moving too far from the draft-conditioned latent basin. These 
are analogous design responses to the same continuous-to-discrete 
problem, operating at different stages of generation.

Latent Diffusion for Language Generation~\cite{lovelace2023latent} 
is especially close to our setting because it trains continuous 
generative models in the latent space of a language autoencoder and 
decodes sampled latent states back into natural language. Our system 
shares this latent-space generative view but differs in three 
important ways. First, we use a flow-based model rather than 
diffusion, updating the full suffix latent sequence jointly in 
parallel. Second, we replace the pretrained autoregressive decoder 
with a parallel decoder trained from scratch, removing left-to-right 
token conditioning during decoding. Third, this setting reveals a 
sharp decoder-readability failure mode: generated latents can be 
geometrically close to real suffix latents while still decoding 
poorly under a frozen parallel decoder. In our experiments, direct 
Gaussian-start latent generation was therefore insufficient, leading 
us to adopt draft-conditioned local refinement, which begins from a 
structured decoder-readable latent state rather than pure noise.

DiffusionBERT~\cite{he2023diffusionbert} provides another relevant 
contrast: it trains BERT directly as the denoising model in a 
discrete diffusion process over token space, making BERT itself the 
generative component. Our system inverts this relationship --- BERT 
is a frozen encoder that defines the latent space, while generation 
and decoding are handled by separate learned components operating in 
continuous space. Where DiffusionBERT asks how diffusion can improve 
masked language model generation, we ask whether continuously 
generated latents can be reliably decoded by a parallel decoder --- 
a question that does not arise in discrete token-space diffusion.

Beyond continuous diffusion, discrete and sequence-level diffusion 
models such as D3PM~\cite{austin2021structured}, 
DiffuSeq~\cite{gong2023diffuseq}, and continuous diffusion for 
categorical data~\cite{dieleman2022continuous} study diffusion-style 
generation over discrete or categorical text spaces. These works 
provide important contrasts to our approach: they modify the text 
generation process itself, whereas we study the failure modes of 
continuous latent refinement followed by parallel decoding.

\subsection{Flow matching and Riemannian flow matching}

Flow Matching~\cite{lipman2023flowmatching}, related to 
rectified-flow and optimal-transport conditional flow 
formulations~\cite{liu2023rectifiedflow,tong2024otcfm}, trains a 
continuous normalizing flow by regressing a neural vector field onto 
a target vector field along a probability path from noise to data:
\begin{equation}
    \mathcal{L}_{\text{FM}} = \mathbb{E}\left[
    \|v_\theta(z_t, t) - u_t\|^2\right], 
    \quad u_t = z_1 - z_0
\end{equation}
Riemannian Flow Matching~\cite{chen2024riemannian} extends this 
idea by defining vector-field matching under a manifold metric. 
This work uses the flow-matching view in a latent text space and 
adds a learned diagonal metric preconditioning. Rather than 
computing true Riemannian geodesics --- which would require 
knowledge of the BERT latent manifold's exponential map and 
connection --- MetricNet learns a diagonal preconditioning matrix 
$G_\psi$ that reweights the force-matching objective in 
decoder-sensitive directions:
\begin{equation}
    \mathcal{L}_{\text{force}} = \mathbb{E}\left[
    \|f_\theta(z_t, t, z^p) - G_\psi \cdot u_t\|^2\right],
    \quad v_{\text{nat}} = G_\psi^{-1} f_\theta
\end{equation}
The diagnostic result is cautionary: metric preconditioning 
improves latent geometry but cannot substitute for a 
decoder-readable start distribution, because the decoder basin 
is a discrete decision boundary that no smooth metric can 
fully approximate.

\subsection{Latent diffusion and structured starts}

Latent Diffusion Models~\cite{rombach2022ldm} show that generative 
modeling can be made more efficient by operating in the latent space 
of an autoencoder. Most closely related to our setting, 
COSMOS~\cite{meshchaninov2025cosmos} addresses latent decodability 
for text diffusion by training the autoencoder jointly for 
token-level reconstruction and alignment with frozen pretrained 
encoder activations, producing a compressed and smoother latent 
space for diffusion. Segment-Level 
Diffusion~\cite{zhu2024sld} addresses long-form text generation 
through segmentation and adversarial latent smoothing; like COSMOS, 
it engineers a more decodable latent space rather than 
characterizing the decodability failure mode directly. Our work is 
complementary to both: rather than primarily proposing a smoothed 
autoencoder space, we empirically characterize the decodability 
failure mode itself, showing that geometric proximity in latent 
space does not necessarily imply token recoverability under a 
parallel decoder. This contrast motivates a further question: 
whether pretrained-encoder alignment alone is sufficient to widen 
the decoder-readable basin for fully non-autoregressive parallel 
decoding, a setting not directly evaluated by COSMOS.

SDEdit~\cite{meng2022sdedit} provides a useful analogy for 
structured-start generation: rather than sampling entirely from 
noise, it starts from a structured guide and denoises toward the 
data manifold. The present system follows a related principle for 
text: a rough continuation draft is encoded into latent space and 
then refined. Unlike SDEdit's continuous image domain, text 
requires the refined latent to land precisely within the 
decoder-readable basin --- the failure mode this work characterizes.

\subsection{Non-autoregressive and iterative generation}

Our work also relates to non-autoregressive and iterative 
generation. Early non-autoregressive translation models generated 
tokens in parallel using latent fertility 
variables~\cite{gu2018nonautoregressive}, while Mask-Predict uses 
conditional masked language modeling for parallel 
refinement~\cite{ghazvininejad2019maskpredict}. Levenshtein 
Transformer~\cite{gu2019levenshtein} and Insertion 
Transformer~\cite{stern2019insertion} further show that sequence 
generation can be framed as iterative editing rather than strict 
left-to-right decoding. Unlike these methods, our system performs 
refinement in a continuous BERT latent space and then evaluates 
whether the resulting states remain decoder-readable.

Text degeneration and decoding choices are also relevant to our 
diagnostics. Prior work shows that decoding strategy can strongly 
affect repetition and 
diversity~\cite{holtzman2020curious,vijayakumar2018diverse}. In our 
setting, argmax collapse is treated as a readout failure rather than 
as direct proof that the latent generator contains no useful 
structure.
\section{Method}

\subsection{Stage-1 latent autoencoder}

Let a token sequence be $\mathbf{x}=(x_1,\ldots,x_n)$ of length 
$n$. A frozen BERT encoder~\cite{devlin2019bert} maps tokens into 
contextual hidden states:
\begin{equation}
    \mathbf{h} = E_{\mathrm{BERT}}(\mathbf{x})
\end{equation}
A projection layer $P$ maps hidden states into a continuous latent 
sequence:
\begin{equation}
    \mathbf{z} = P(\mathbf{h}), \quad \mathbf{z} \in 
    \mathbb{R}^{n \times d}
\end{equation}
where $d$ is the latent dimension ($d=768$ in our main experiments, 
$d=256$ for the compressed baseline). A parallel decoder $p_\phi$ 
predicts all token distributions simultaneously:
\begin{equation}
    p_\phi(x_i \mid \mathbf{z}) = 
    \mathrm{softmax}(D_\phi(\mathbf{z})_i)
\end{equation}
The reconstruction objective is:
\begin{equation}
    \mathcal{L}_{\mathrm{AE}} = -\sum_{i=1}^{n} 
    \log p_\phi(x_i \mid \mathbf{z})
\end{equation}
We evaluate both compressed 256-dimensional latents and full 
768-dimensional BERT latents. The 768-dimensional setting removes 
the projection bottleneck as much as possible while keeping the 
same decoder interface.

\subsection{Conditional latent flow matching}

The latent sequence is split into prompt and suffix parts at 
position $m$:
\begin{equation}
    \mathbf{z} = (\mathbf{z}^p, \mathbf{z}^s), \quad
    \mathbf{z}^p = \mathbf{z}_{1:m}, \quad 
    \mathbf{z}^s = \mathbf{z}_{m+1:n}
\end{equation}
where $\mathbf{z}^p \in \mathbb{R}^{m \times d}$ is the known 
prompt latent and $\mathbf{z}^s \in \mathbb{R}^{(n-m) \times d}$ 
is the suffix latent to be generated. In our ROCStories 
experiments, $m=32$ prompt token slots and $n-m=32$ suffix token 
slots.

Following the flow-matching view~\cite{lipman2023flowmatching}, 
in the original Gaussian-start formulation, a path interpolates 
between noise and the real suffix latent:
\begin{equation}
    \mathbf{z}_0 \sim \mathcal{N}(0, I), \quad 
    \mathbf{z}_1 = \mathbf{z}^s
\end{equation}
\begin{equation}
    \mathbf{z}_t = (1-t)\mathbf{z}_0 + t\mathbf{z}_1, \quad 
    t \sim \mathcal{U}(0,1)
\end{equation}
where $I$ is the identity matrix and $t$ is the interpolation 
timestep. The target velocity is:
\begin{equation}
    \mathbf{u}_t = \mathbf{z}_1 - \mathbf{z}_0
\end{equation}
FlowNet $v_\theta$ predicts a conditional velocity field, and the 
basic flow-matching objective is:
\begin{equation}
    \mathcal{L}_{\mathrm{FM}} = \mathbb{E}\left[
    \left\|v_\theta(\mathbf{z}_t, t, \mathbf{z}^p) - 
    (\mathbf{z}_1 - \mathbf{z}_0)\right\|_2^2\right]
\end{equation}
Diagnostics showed this Gaussian-start formulation was 
insufficient: generated latents $\hat{\mathbf{z}}^s$ could 
achieve high cosine similarity with real suffix latents while 
still decoding poorly:
\begin{equation}
    \cos(\mathbf{z}^s, \hat{\mathbf{z}}^s) \approx 1 
    \not\Rightarrow P_{\text{target}}(\hat{\mathbf{z}}^s) 
    \text{ is high}
\end{equation}
where $P_{\text{target}}$ is defined formally in Eq.~\ref{eq:ptarget}.

\subsection{Metric-aware local refinement}

Rather than computing true Riemannian geodesics --- which would 
require knowledge of the BERT latent manifold's exponential map 
and connection --- MetricNet learns a diagonal preconditioning 
matrix $G_\psi$ that reweights the force-matching objective in 
decoder-sensitive directions:
\begin{equation}
    G_\psi(\mathbf{z}_t, t, \mathbf{z}^p) \succ 0, \quad 
    G_\psi \in \mathbb{R}^{d \times d} \text{ diagonal}
\end{equation}
FlowNet is trained to predict a force/covector 
$f_\theta \approx G_\psi \mathbf{u}_t$ rather than the velocity 
directly. The force target is:
\begin{equation}
    f_{\text{target}} = G_\psi(\mathbf{z}_t, t, \mathbf{z}^p) 
    \cdot \mathbf{u}_t
\end{equation}
and the force-matching objective is:
\begin{equation}
    \mathcal{L}_{\text{force}} = \mathbb{E}\left[
    \left\|f_\theta(\mathbf{z}_t, t, \mathbf{z}^p) - 
    G_\psi(\mathbf{z}_t, t, \mathbf{z}^p) \cdot 
    \mathbf{u}_t\right\|^2\right]
\end{equation}
At inference, natural velocity is recovered by inverting the 
metric:
\begin{equation}
    v_{\text{nat}} = G_\psi^{-1} f_\theta = 
    \frac{f_\theta}{g_\psi}
\end{equation}
and the ODE update becomes:
\begin{equation}
    \mathbf{z}_{t+\Delta t} = \mathbf{z}_t + 
    \gamma \cdot v_{\text{nat}}(\mathbf{z}_t, t, \mathbf{z}^p) 
    \cdot \Delta t, \quad \gamma = 0.01
\end{equation}
A small regularizer keeps the metric near identity early in 
training:
\begin{equation}
    \mathcal{L}_{\mathrm{metric}} = \|G_\psi - I\|_F^2
\end{equation}
In the current implementation, MetricNet is a small MLP with 
input dimension $2d+2$, hidden dimension 256, and output 
dimension $d$. The metric log values are bounded by 0.50, then 
exponentiated and normalized. For ROCStories with $d=768$, 
MetricNet has 0.66M parameters. Later runs showed that MetricNet 
can learn nontrivial anisotropy, for example diagonal standard 
deviation around 0.38 and range approximately $[0.47, 1.28]$. 
However, this geometric improvement only gives modest 
decoder-level gains, confirming that metric preconditioning 
alone cannot substitute for a decoder-readable start 
distribution.

\subsection{Draft-conditioned start distribution}

Prompt-only generation is highly multimodal. One prompt can have 
many valid continuations, so a prompt-only latent prior can 
average over incompatible targets. We therefore use a rough draft 
$r$ as an instance-level guide. The draft is encoded by the same 
frozen BERT encoder:
\begin{equation}
    \mathbf{z}_{\text{draft}} = P(E_{\text{BERT}}(r))
\end{equation}
The DraftPrior is implemented as a denoising prior over suffix 
latents. The input draft latent is mixed with Gaussian noise:
\begin{equation}
    \mathbf{z}_t = \alpha \mathbf{z}_{\text{draft}} + 
    \sqrt{1-\alpha^2}\,\epsilon, \quad 
    \alpha = 0.7, \quad 
    \epsilon \sim \mathcal{N}(0, I)
\end{equation}
The model predicts a residual correction:
\begin{equation}
    \mathbf{z}_{\text{start}} = \mathbf{z}_t + 
    f_\eta(\mathbf{z}_t, \mathbf{z}^p, \alpha)
\end{equation}
where $f_\eta$ is the DraftPrior network and $\mathbf{z}^p$ is 
the prompt latent. It is trained with a combined objective:
\begin{equation}
    \mathcal{L}_{\text{DP}} = \mathcal{L}_{\text{CE}} + 
    \lambda_1 \mathcal{L}_{\text{MSE}} + 
    \lambda_2 \mathcal{L}_{\text{cos}} + 
    \lambda_3 \mathcal{L}_{\text{norm}}
\end{equation}
where the regularizers are defined as:
\begin{align}
    \mathcal{L}_{\text{MSE}} &= 
    \|\mathbf{z}_{\text{start}} - \mathbf{z}^s\|^2 \\
    \mathcal{L}_{\text{cos}} &= 
    1 - \cos(\mathbf{z}_{\text{start}}, \mathbf{z}^s) \\
    \mathcal{L}_{\text{norm}} &= 
    \left(\|\mathbf{z}_{\text{start}}\|_2 - 
    \|\mathbf{z}^s\|_2\right)^2
\end{align}
The norm regularizer explicitly trains DraftPrior to match both 
the angle and magnitude of the real suffix latent --- addressing 
the limitation of cosine similarity alone, which measures only 
angular agreement.

Architecturally, the DraftPrior uses an $\alpha$ embedding, a 
position embedding, and four Transformer-style layers. Each layer 
has suffix self-attention, prompt cross-attention, and a 
feed-forward block. The model uses eight attention heads and a 
feed-forward hidden dimension of 512.

Draft corruption is simple: target tokens are dropped with 
probability 0.05 and are not replaced. This creates a noisy but 
structured draft. The goal is not prompt-only generation; the 
goal is to test whether the model can refine a decoder-readable 
but imperfect latent start.

The local FlowNet then refines $\mathbf{z}_{\text{start}}$ 
rather than generating from pure Gaussian noise. The local 
refinement target is a small residual fraction of the full 
displacement:
\begin{equation}
    \mathbf{u}_{\text{local}} = 
    \rho(\mathbf{z}^s - \mathbf{z}_{\text{start}}), 
    \quad \rho = 0.05
\end{equation}
The ODE update uses a small scalar $\gamma = 0.01$ that prevents 
the flow from moving far enough to destroy the draft-conditioned 
decoder-readable structure.

\subsection{Auxiliary fused readout, OT regularization, and 
residual refinement}

Because raw latent movement often changes decoder CE only weakly, 
we also evaluate an auxiliary token head and fused readout:
\begin{equation}
    \ell_{\text{fused}} = \ell_{\text{decoder}} + 
    \beta \ell_{\text{aux}}, \quad \beta = 0.1
\end{equation}
where the two loss terms are:
\begin{align}
    \ell_{\text{decoder}} &= -\sum_{i \in M} 
    \log p_\phi(x_i \mid \mathbf{z}_T) \\
    \ell_{\text{aux}} &= -\sum_{i \in M} 
    \log p_{\text{aux}}(x_i \mid \mathbf{z}_t)
\end{align}
and $M$ is the set of valid target token positions. This readout 
is decoder-aware and is often more effective than moving the 
latent alone.

We additionally tested distribution-level alignment between 
refined suffix latents and real suffix latents using an 
optimal-transport-style loss:
\begin{equation}
    \mathcal{L}_{\mathrm{OT}} = 
    \mathrm{OT}(\hat{\mathbf{z}}^s, \mathbf{z}^s)
\end{equation}
implemented with Sinkhorn OT when available and a 
sliced-Wasserstein-style fallback otherwise. Finally, we tested 
bounded residual refinement as an ablation:
\begin{equation}
    \mathbf{z}_{\text{res}} = \mathbf{z}_{\text{ode}} + 
    \lambda \tanh\left(R_\omega(\mathbf{z}_{\text{ode}}, 
    \mathbf{z}^p)\right)
\end{equation}
where $\lambda > 0$ bounds the maximum refinement magnitude, 
ensuring $\|\mathbf{z}_{\text{res}} - 
\mathbf{z}_{\text{ode}}\|_\infty \leq \lambda$. These extensions 
are treated as ablations, not as the main result.

\subsection{Full inference procedure}
\label{sec:inference}

Algorithm~\ref{alg:inference} summarizes the complete 
draft-conditioned latent refinement pipeline at inference time.

\begin{algorithm}[h]
\caption{Draft-Conditioned Latent Refinement}
\label{alg:inference}
\begin{algorithmic}[1]
\Require prompt $x^p$, draft $r$, ODE steps $T$, 
         scale $\gamma=0.01$
\State $\mathbf{z}^p \leftarrow P(E_{\text{BERT}}(x^p))$
    \Comment{Encode prompt}
\State $\mathbf{z}_{\text{draft}} \leftarrow 
    P(E_{\text{BERT}}(r))$
    \Comment{Encode draft}
\State $\mathbf{z}_0 \leftarrow 
    \text{DraftPrior}(\mathbf{z}_{\text{draft}}, \mathbf{z}^p)$
    \Comment{Decoder-readable start}
\For{$t = 0, \Delta t, 2\Delta t, \ldots, 1$}
    \State $f_\theta \leftarrow 
        \text{FlowNet}(\mathbf{z}_t, t, \mathbf{z}^p)$
        \Comment{Predict force}
    \State $v_{\text{nat}} \leftarrow 
        f_\theta \,/\, 
        G_\psi(\mathbf{z}_t, t, \mathbf{z}^p)$
        \Comment{Natural velocity}
    \State $\mathbf{z}_{t+\Delta t} \leftarrow 
        \mathbf{z}_t + \gamma \cdot v_{\text{nat}} \cdot \Delta t$
        \Comment{ODE step}
\EndFor
\State $\hat{x} \leftarrow 
    \arg\max\, p_\phi(\cdot \mid \mathbf{z}_T)$
    \Comment{Parallel decode}
\Ensure token sequence $\hat{x}$
\end{algorithmic}
\end{algorithm}

\section{Experimental Setup}

\subsection{Implementation Details}

All experiments were conducted on a single NVIDIA RTX 4090 GPU with 24GB VRAM using PyTorch. Unless otherwise specified, models were trained with a fixed random seed of 1337. We report ROCStories experiments with full 768-dimensional BERT latents and a compressed 256-dimensional diagnostic baseline, maximum sequence length of 64, batch size up to 768, and up to 16-step ODE integration during inference. The implementation was maintained with a chronological change log recording architectural switches, evaluation fixes, and benchmark protocol changes; this was especially important because several early rows were diagnostics rather than final comparable benchmark results.

Table~\ref{tab:params} reports the main trainable component sizes for the compressed 256-dimensional configuration and the full 768-dimensional ROCStories configuration. FlowNet uses hidden dimension 512, depth 5, eight attention heads, depthwise convolution with kernel size 5, suffix self-attention, prompt cross-attention, and residual output heads. The ROCStories first2/last3 split uses 32 prompt slots and 32 target slots.

\begin{table}[h]
\centering
\caption{Trainable component sizes in the current implementation.}
\label{tab:params}
\renewcommand{\arraystretch}{1.15}
\begin{tabular}{lrr}
\toprule
Component & Compressed 256-d & ROCStories 768-d \\
\midrule
DraftPrior & 3.36M & 23.84M \\
FlowNet & 15.93M & 27.74M \\
MetricNet & 0.26M & 0.66M \\
\bottomrule
\end{tabular}
\end{table}

\subsection{Dataset and split}

We evaluate on ROCStories~\cite{mostafazadeh2016rocstories}, where each example is a five-sentence commonsense story. We use a sentence-aware split:
\[
  \text{prompt}=\text{sentences 1--2},\qquad
  \text{target}=\text{sentences 3--5}.
\]
The prompt and target are each allocated 32 BERT-token slots, giving a maximum length of 64 tokens. This avoids cutting prompts in the middle of sentences and better matches story continuation.

\subsection{Draft corruption}

For controlled refinement, the target continuation is corrupted to form a rough draft. The primary corruption is token dropout with no replacement. This setting is not prompt-only generation. It measures whether the model can refine a noisy structured draft back toward a decoder-readable continuation. We report 0\%, 3\%, 5\%, and 10\% dropout settings.

\subsection{Metrics}
\label{sec:metrics}

For decoder recoverability we report cross-entropy (CE), average target-token probability $p$, and top-1 token accuracy under the frozen decoder. For text-level comparisons we report MAUVE~\cite{pillutla2021mauve}, distinct-1/2, repetition rate, average generated length, latency per sample, and tokens per second. Diffusion-LM is discussed only as a literature reference unless explicitly reproduced under the same local ROCStories export, truncation, and scoring protocol. The reported Diffusion-LM MAUVE value is therefore not an apples-to-apples row in our benchmark tables. Because MAUVE is unstable with small sample counts and short continuations, the 500-sample MAUVE results should be interpreted as preliminary.

\section{Results}

\subsection{DraftPrior corruption curve}

Table~\ref{tab:draftprior} shows the main DraftPrior corruption curve. Moving from compressed 256-dimensional latents to full 768-dimensional BERT latents consistently improves recoverability. The improvement is not proportional to dimensionality; rather, it suggests that compression removes specific token-recovery information such as subword identity, rare words, and lexical choice.

\begin{table}[h]
\centering
\caption{DraftPrior recovery on ROCStories under the first2/last3 split. Values are validation metrics under the frozen decoder.}
\label{tab:draftprior}
\renewcommand{\arraystretch}{1.15}
\begin{tabular}{lrrrr}
\toprule
Setting & Dropout & CE $\downarrow$ & Target prob. $\uparrow$ & Top-1 $\uparrow$ \\
\midrule
256-d latent & 0\% & 0.39 & 0.87 & 0.89 \\
256-d latent & 3\% & 2.608 & 0.543 & 0.627 \\
256-d latent & 5\% & 3.165 & 0.434 & 0.525 \\
768-d latent & 0\% & \textbf{0.235} & \textbf{0.938} & \textbf{0.944} \\
768-d latent & 3\% & \textbf{2.228} & \textbf{0.613} & \textbf{0.677} \\
768-d latent & 5\% & \textbf{3.069} & \textbf{0.483} & \textbf{0.573} \\
768-d latent & 10\% & 4.112 & 0.272 & 0.357 \\
\bottomrule
\end{tabular}
\end{table}

The clean 768-dimensional setting is nearly decoder-readable. Mild corruption remains usable: 3\% dropout reaches target probability 0.613 and 5\% dropout reaches 0.483. At 10\% dropout, performance drops sharply and qualitative samples show tail repetition and function-word loops. We therefore treat 10\% as a stress-test regime rather than the main operating point.

\subsection{Stage-2 local refinement}

\begin{figure}[H]
\centering
\includegraphics[width=\linewidth]{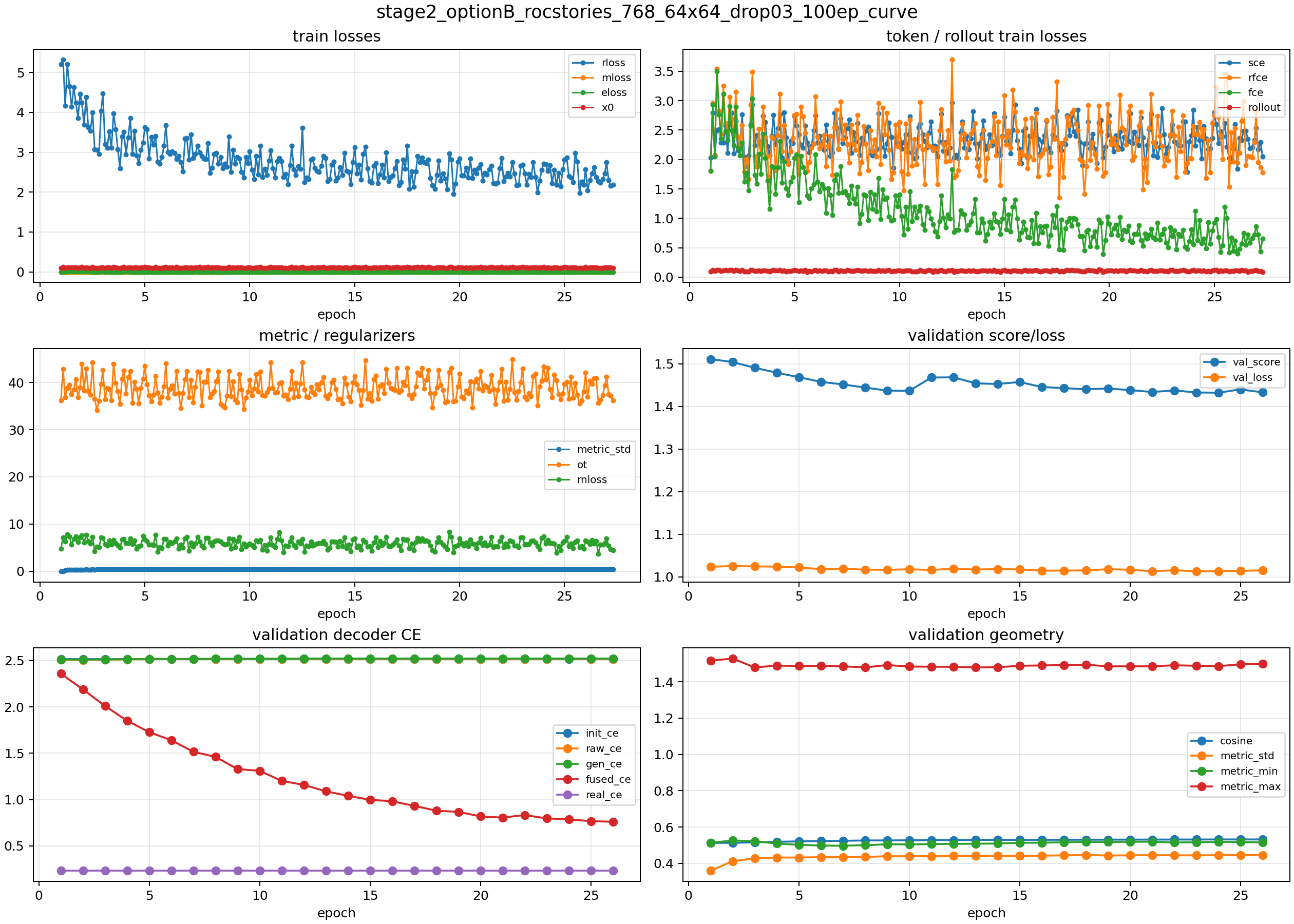}
\caption{Stage-2 training dynamics for the ROCStories 768-dimensional 64/64 run with 3\% draft dropout. The curve shows that decoder recoverability improves strongly for the fused readout, while the raw generated CE remains much worse than the real-latent decoder CE. This supports the paper's main claim that local geometry and decoder-readable token recovery must be evaluated separately.}
\label{fig:stage2_curve}
\end{figure}

Table~\ref{tab:stage2} summarizes representative 768-dimensional stage-2 validation snapshots using the 5\% DraftPrior start. DraftPrior provides most recoverability. Raw local flow makes only small changes, while fused decoder-aware readout gives the clearest improvement. Scaling MetricNet and adding OT makes the metric nontrivial, but the decoder-level gain remains modest. A bounded residual refiner moves latents more, but did not improve the best validation recoverability in these runs. Figure~\ref{fig:stage2_curve} shows the corresponding training dynamics for a 3\% draft-dropout 768-dimensional run.

\begin{table}[h]
\centering
\caption{Representative 768-dimensional stage-2 refinement snapshots with a 5\% DraftPrior start. Values come from validation diagnostics.}
\label{tab:stage2}
\renewcommand{\arraystretch}{1.15}
\begin{tabular}{lrrr}
\toprule
Stage / variant & CE $\downarrow$ & Target prob. $\uparrow$ & Note \\
\midrule
DraftPrior start & 3.193 & 0.457 & Structured start \\
Raw local flow & 3.219 & $\approx$0.46 & Little direct gain \\
Fused readout & \textbf{2.904} & \textbf{0.493} & Best decoder-aware snapshot \\
Active MetricNet + OT & 3.057 & 0.482 & Nontrivial metric, modest gain \\
Bounded residual ablation & 3.120 & 0.472 & More movement, no clear gain \\
Oracle real latents & 0.172 & 0.953 & Frozen-decoder upper bound \\
\bottomrule
\end{tabular}
\end{table}

The gap to the oracle remains large. This shows that the remaining bottleneck is not merely latent distribution matching, but precise placement inside the decoder-readable basin.

\subsection{Interpolation diagnostic}

The interpolation diagnostic in Table~\ref{tab:mixce} moves generated latents toward real latents:
\[
  \mathbf{z}_{\alpha}=(1-\alpha)\hat{\mathbf{z}}+\alpha\mathbf{z}^{s}.
\]
Small movement toward the real latent sharply improves CE, confirming that generated latents are in the right neighborhood but still outside the sharp decoder basin.

\begin{table}[h]
\centering
\caption{Interpolation diagnostic for 768-dimensional generated latents.}
\label{tab:mixce}
\renewcommand{\arraystretch}{1.15}
\begin{tabular}{lrrrrrr}
\toprule
$\alpha$ & 0.01 & 0.03 & 0.05 & 0.10 & 0.20 & 0.50 \\
\midrule
CE $\downarrow$ & 3.188 & 3.123 & 3.057 & 2.885 & 2.482 & 0.589 \\
\bottomrule
\end{tabular}
\end{table}

\subsection{Quality-speed curve}

Table~\ref{tab:curve} reports a preliminary 500-sample ROCStories quality-speed curve for the full 768-dimensional system using 64 fixed BERT suffix slots. The zero-step row corresponds to the DraftPrior-only output before ODE refinement. All rows in the table were produced by the local benchmark. The commonly cited Diffusion-LM MAUVE value of 0.043 is a reported literature number, not a result from this local ROCStories benchmark, so it is excluded from the table and used only as background context.

\begin{table}[h]
\centering
\caption{Preliminary 500-sample ROCStories quality-speed curve from the local benchmark only.}
\label{tab:curve}
\renewcommand{\arraystretch}{1.15}
\begin{tabular}{lrrrrr}
\toprule
Model & Steps & MAUVE $\uparrow$ & Dist-1 $\uparrow$ & Rep. $\downarrow$ & Latency $\downarrow$ \\
\midrule
Ours 768 & 0 & 0.0075 & 0.0323 & 0.0130 & 0.0034s \\
Ours 768 & 1 & 0.0062 & 0.0310 & 0.0123 & 0.0032s \\
Ours 768 & 2 & 0.0075 & 0.0316 & 0.0127 & 0.0033s \\
Ours 768 & 4 & 0.0068 & 0.0315 & 0.0121 & 0.0036s \\
Ours 768 & 8 & 0.0069 & 0.0311 & 0.0126 & 0.0040s \\
Ours 768 & 16 & 0.0074 & 0.0314 & 0.0111 & 0.0050s \\
\bottomrule
\end{tabular}
\end{table}

A separate sentence-split local comparison gives GPT-2 autoregressive MAUVE 0.0067 and the 16-step 768-dimensional draft-conditioned system MAUVE 0.0109 under the same exported ROCStories rows. This comparison should still be read cautiously because GPT-2 is prompt-only autoregressive generation, while our system is synthetic-draft-conditioned latent refinement.

These results are not a state-of-the-art claim. They show that the controlled synthetic-draft system is not random, but its absolute MAUVE remains low. The main value of the report is the diagnostic analysis of recoverability, not benchmark dominance.

\subsection{Qualitative behavior}

Table~\ref{tab:qualitative} shows representative samples across three behavioral regimes. Near-perfect recovery occurs at short-to-medium target lengths (15--25 tokens): the system recovers event structure and most lexical content, with errors limited to dropped leading tokens or tail artifacts. Partial recovery at medium lengths (25--32 tokens) shows semantic near-synonyms replacing target words (``toppled''$\to$``tumbled'', ``pyramid''$\to$``channel'', ``faulty''$\to$``failure'') and entity substitution (``conan''$\to$``mickey''/``tom''). Full collapse occurs at longer targets ($\geq$32 tokens), where output degenerates into function-word loops. The 256-dimensional condition shows a qualitatively different failure mode: event scaffold is preserved but key nouns are wrong (``wide receiver''$\to$``brand wave'', ``play''$\to$``class''), consistent with compression removing lexical specificity while retaining syntactic structure.

\begin{table}[h]
\centering
\caption{Selected qualitative samples. \textbf{Ref} is the ground-truth continuation; \textbf{Pred} is the system output. Substitution errors are shown in \textit{italics}; dropped tokens in [\,]. $L$ = target length in tokens.}
\label{tab:qualitative}
\renewcommand{\arraystretch}{1.25}
\begin{tabular}{p{3.0cm} p{4.6cm} p{4.6cm} r}
\toprule
Prompt & Ref & Pred & $L$ \\
\midrule
\multicolumn{4}{l}{\textit{Near-perfect recovery (768-d)}} \\
\midrule
Joseph clipped his toe nails. They went flying across the room. &
one hit his sister on the head. his sister was disgusted. she slapped him and pushed him out of the room. &
\textit{[one]} hit his sister on the head. his sister was disgusted. she slapped him and pushed him out of the room.\textit{,} &
24 \\
\addlinespace
Ann went to a movies. She took a seat in a front row. &
her neighbor accidentally spit some coke on her pants. he apologized. they started a nice conversation. &
\textit{[her]} neighbor accidentally spit some coke on her pants. he apologized. they started a nice conversation. \textit{got} &
19 \\
\midrule
\multicolumn{4}{l}{\textit{Partial recovery --- lexical substitution (768-d)}} \\
\midrule
Kasey was a cheerleader. She was performing at a football game. &
she climbed to the top of the cheer pyramid. but then she toppled to the ground! kasey was lucky not to be injured. &
climbed to the top of the cheer \textit{channel}. but then she \textit{tumbled} to the ground! kasey was lucky not to be \textit{bruised} and &
26 \\
\addlinespace
Conan was a bright red headed man. He wanted to host a show. &
the network decided it was time for him to get his own show. conan was ready to take on the role. conan loved his new job. &
network decided it was time for to get his own show. \textit{mickey} was ready to take on the role. \textit{tom} loved. new job and show... &
31 \\
\midrule
\multicolumn{4}{l}{\textit{Collapse --- long target (768-d)}} \\
\midrule
Jeff was at a red light. All of a sudden it turned green. &
as he was getting ready to accelerate, a car bumped into him. he got ready to pull over to exchange information and call the cops. &
he was getting ready to fast a car bumped an him as. got had to taking while the the a to a was.,. a he out t. \ldots &
41 \\
\midrule
\multicolumn{4}{l}{\textit{Compression failure (256-d)}} \\
\midrule
Josh was the quarterback. His team was down by six points. &
on the last play, a wide receiver got open in the end zone. josh threw the ball but it went over the receiver's head. josh felt bad. &
the last \textit{class}, a \textit{brand wave} got open in the end zone. josh threw the ball but it went over the receiver's head. josh bad for letting his team down. &
36 \\
\bottomrule
\end{tabular}
\end{table}

\section{Failure Mode Analysis}

\subsection{Latent similarity is not discrete recovery}

The central failure mode is that continuous similarity does not imply discrete recoverability. A generated suffix latent can be close to the real suffix latent under cosine similarity or MSE while still lying outside the decoder decision region that produces the correct token:
\[
  \text{high }\cos(\mathbf{z}^s,\hat{\mathbf{z}}^s)
  \not\Rightarrow
  \text{high }P_{\text{target}}.
\]
For a generated suffix latent $\hat{\mathbf{z}}^s$, define target-token probability as
\begin{equation}
    P_{\text{target}}(\hat{\mathbf{z}}^s) = \frac{1}{|\mathcal{M}|}
    \sum_{i \in \mathcal{M}} p_\phi(x_i \mid \hat{\mathbf{z}}^s)
    \label{eq:ptarget}
\end{equation}
where $\mathcal{M}$ is the valid target-token mask. This recoverability metric is more informative than latent cosine alone.

\subsection{Compression removes token-recovery information nonlinearly}

The 768-dimensional results show that compression is not neutral. Compressing BERT~\cite{devlin2019bert} contextual states to 256 dimensions preserves some semantic scaffold but loses token-level information needed for exact recovery. This explains errors such as semantically plausible but lexically wrong substitutions. The improvement from 256 to 768 is not $3\times$, but it is consistent across corruption levels.

\subsection{Argmax collapse is a readout failure}

When decoder logits are broad, argmax repeatedly selects tokens with slightly larger output bias, often punctuation, articles, or common function words. This can create the appearance of total collapse. Sampling from the same logits may reveal more structure. Argmax collapse should therefore be reported as a diagnostic, not as the only evaluation of the latent generator.

\subsection{Moving more is not enough}

Once the DraftPrior start is decoder-readable, large flow updates can damage it. The residual ablation confirms that simply adding more latent movement is not sufficient. The movement must be decoder-aligned. Similarly, OT regularization can make latent distributions more geometrically aligned without guaranteeing token recovery. This reinforces the main thesis: geometry helps only when it respects the decoder interface.

\section{Limitations}

This is a diagnostic technical report, not a claim that the system solves non-autoregressive text generation. The strongest current result is controlled draft-conditioned refinement. Synthetic corrupted drafts contain target-derived information and must not be described as prompt-only generation. A mature comparison still requires stronger baselines such as discrete diffusion~\cite{austin2021structured}, DiffuSeq~\cite{gong2023diffuseq}, Mask-Predict~\cite{ghazvininejad2019maskpredict}, Levenshtein Transformer~\cite{gu2019levenshtein}, insertion-based models~\cite{stern2019insertion}, VQ/codebook latents, and externally generated drafts.

The Riemannian component should also be interpreted carefully. Larger MetricNet variants can learn nontrivial metric anisotropy, but current metric and OT experiments produce only modest decoder-level gains. The present claim is not that learned Riemannian geometry solves text generation. The claim is that decoder-readable starts are necessary before geometry can help.

Finally, fixed-length target slots can create tail artifacts after the semantic continuation has ended. Future evaluation should truncate at an end marker where possible and compute token metrics only over valid target positions.

\section{Conclusion}

This report began with a simple question: can conditional flow matching in BERT latent space generate text continuations non-autoregressively? The diagnostic answer is more subtle than yes or no. Pure Gaussian latent generation is too weak because it lacks instance-level token information and often falls outside the decoder-readable basin. Prompt-only priors also struggle because valid continuations are multimodal. Structured draft conditioning changes the problem: it provides a readable start latent, after which FlowNet and MetricNet can be treated as local refiners. Full 768-dimensional BERT latents improve recovery over compressed 256-dimensional latents, confirming that compression can remove token-level information needed for discrete decoding. The main lesson is that latent geometry is not enough. Continuous-latent text generation must be evaluated by discrete recoverability, start-distribution quality, and whether refinement preserves rather than destroys decoder-readable structure.

\medskip
\noindent\textit{Stage 1 status.} The latent autoencoder, DraftPrior, and recoverability diagnostics reported here are complete. Stage 2---stronger refinement models, externally generated drafts, and broader baselines---remains ongoing.

\bibliographystyle{plain}

\end{document}